\documentclass[10pt,twocolumn,letterpaper]{article}

\usepackage{cvpr}              

\usepackage{graphicx}
\usepackage{amsmath}
\usepackage{amssymb}
\usepackage{mathrsfs}
\usepackage{booktabs}
\usepackage{algorithm}
\usepackage{algpseudocode}
\usepackage{makecell}
\usepackage{multirow}
\usepackage{enumitem}

\definecolor{cvprblue}{rgb}{0.21,0.49,0.74}
\usepackage[pagebackref,breaklinks,colorlinks,allcolors=cvprblue]{hyperref}

\begin{document}

\title{HV-Attack: Hierarchical Visual Attack for Multimodal Retrieval\\ Augmented Generation}

\author{Linyin Luo$^{1,2}$, Yujuan Ding$^1$, Yunshan Ma$^3$, Wenqi Fan$^1$, Hanjiang Lai$^2$\\
$^1$The Hong Kong Polytechnic University\\
$^2$Sun Yat-Sen University\\
$^3$Singapore Management University\\
{\tt\small luoly36@mail2.sysu.edu.cn, dingyujuan385@gmail.com, ysma\@smu.edu.sg}\\
{\tt\small wenqifan03@gmail.com, laihanj3@mail.sysu.edu.cn}
}

\maketitle

\begin{abstract}
Advanced multimodal Retrieval-Augmented Generation (MRAG) techniques have been widely applied to enhance the capabilities of Large Multimodal Models (LMMs), but they also bring along novel safety issues. Existing adversarial research has revealed the vulnerability of MRAG systems to knowledge poisoning attacks, which fool the retriever into recalling injected poisoned contents. However, our work considers a different setting: \textbf{visual attack of MRAG by solely adding imperceptible perturbations at the image inputs of users, without manipulating any other components.} This is challenging due to the robustness of fine-tuned retrievers and large-scale generators, and the effect of visual perturbation may be further weakened by propagation through the RAG chain. We propose a novel Hierarchical Visual Attack that misaligns and disrupts the two inputs (the multimodal query and the augmented knowledge) of MRAG's generator to confuse its generation. We further design a hierarchical two-stage strategy to obtain misaligned augmented knowledge. We disrupt the image input of the retriever to make it recall irrelevant knowledge from the original database, by optimizing the perturbation which first breaks the cross-modal alignment and then disrupts the multimodal semantic alignment. 
We conduct extensive experiments on two widely-used MRAG datasets: OK-VQA and InfoSeek. We use CLIP-based retrievers and two LMMs BLIP-2 and LLaVA as generators. Results demonstrate the effectiveness of our visual attack on MRAG through the significant decrease in both retrieval and generation performance.
\end{abstract}    
\section{Introduction}
\label{sec:intro}
\begin{figure}[t]
  \centering
   \includegraphics[width=\linewidth]{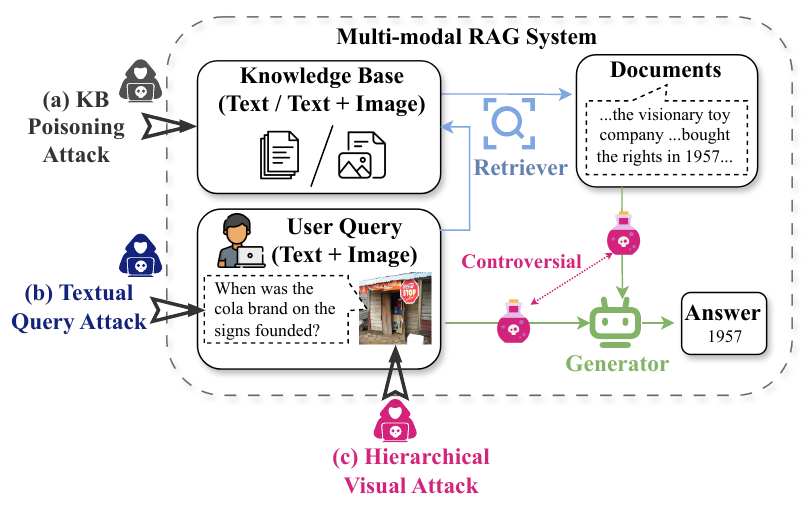}

   \caption{Comparison on (a) KB Poisoning Attack, (b) Textual Query Attack and our focused task (c) Hierarchical Visual Attack on multi-modal RAG.}
   \label{fig:task}
\end{figure}
The robustness of multimodal Retrieval-Augmented Generation (MRAG)~\cite{mei2025mrag-survey, abootorabi2025ask, hu2024mrag, xia2024mmed} systems is of great importance. As advanced MRAG techniques have been widely applied to large multimodal models (LMMs)~\cite{liu2023visual,li2023blip,yang2025qwen3, liu2024gpt} to enhance their knowledge and ability, novel safety issues have also emerged. Existing adversarial research has revealed the vulnerability of MRAG to knowledge poisoning attacks~\cite{liu2025poisoned, zou2025poisonedrag, ha2025mmpoison}. These methods rely on constructing and injecting malicious contents into the knowledge base, which are obvious by changing textual words or can be identified by Knowledge Base (KB) detection algorithm. Meanwhile, there are research of attacks on the generator component of MRAG~\cite{schlarmann2023adversarial, yin2024vqattack, wang2024transferable}, which include manipulations on textual queries and are easy to notice as well. Thus, in this paper, we consider a different setting of attacking MRAG: pure visual attack that solely adds imperceptible perturbations at the image inputs of users. As shown in \autoref{fig:task}, our attack method aims to disrupt both retrieval and generation process of MRAG by optimizing visual perturbations, without manipulating any other components. Compared to poisoning attacks, adversarial visual attack on images can be highly imperceptible to human vision, making them exceptionally stealthy.

\begin{figure}
    \centering
    \includegraphics[width=1\linewidth]{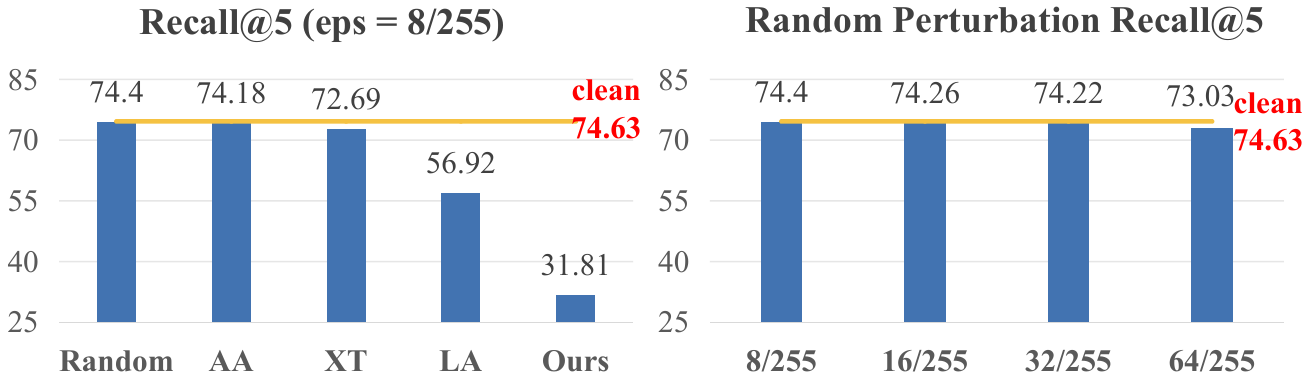}
    \caption{Attack performance on a CLIP model fine-tuned for the retrieval task by different attack methods (AA: Any Attack~\cite{zhang2025anyattack}; XT: X-Transfer~\cite{huang2025xtransfer}, LA: LMM Attack~\cite{cui2024robustness}) and scale. }
    \label{fig:attack_challenge}
\end{figure}

However, attacking MRAG with pure visual perturbation is challenging. The MRAG system is a pipeline that consists of a fine-tuned retriever and a large-scale generator, both of which retain a certain degree of robustness. For fine-tuned retrievers, as shown in \autoref{fig:attack_challenge}, advanced visual attack methods~\cite{zhang2025anyattack, huang2025xtransfer, cui2024robustness} can cause limited retrieval performance drop, and random perturbations even with a large scale (64/255) causes subtle performance drop. This may be because the models become adept at multimodal knowledge grounding and logical association, focusing on relational understanding (e.g., the logical relationship between a question and a potential answer). This emphasis on deeper logical correlation and inference ability makes the model less susceptible to simple input perturbations in a single modality, significantly increasing the challenge of our image attack task. For generators, large-scale LMMs such as LLaVA~\cite{liu2023visual} and BLIP-2~\cite{li2023blip} are often used, which 
generally demonstrate superior robustness~\cite{zeng2025rare}, further posing challenging to MRAG attack. 

The sequential working pipeline of MRAG also enhances its robustness towards visual perturbations as the attack effect degrades by propagation through the RAG chain. The perturbation's effect on retriever lies in the recalled knowledge, which experiences chain of transformation and decrease. Consequently, success visual attack on MRAG is hard, while posing a more severe threat than obvious text attacks or KB poisoning attacks, as the resulting bad influence can go undetected and propagate through the entire RAG chain. 

To effectively subvert the robustness of MRAG with only visual perturbations, we propose a novel hierarchical disruption approach to attack the MRAG's on two components and mislead the final result. As shown in \autoref{fig:task}, there are two inputs for the MRAG generator: the multimodal user query and the augmented knowledge. Our hierarchical method targets at misaligning and disrupting these two inputs, creating knowledge conflicts for the generator while breaking different levels of model capabilities. For retrieval, we employ a hierarchical two-stage strategy to optimize the applied noise in modality and semantic levels. We first alter the query features to no longer correspond accurately to itself, so that the retrieval query is deviated. Then, the semantic alignment between query and knowledge is broken down. 
By structurally targeting the model's core competencies across different levels of abstraction, our approach achieves severe and effective degradation regarding retrieval and generation performance.

In summary, our main contributions are as follows:
\begin{itemize}
    \item We propose a novel Hierarchical Visual Attack method, which misaligns and disrupts the two inputs of generator in MRAG, posting stealth and severe threats for MRAG systems.
    \item Our method is the first to disrupt MRAG systems only using image perturbations, which further reveals the vulnerabilities of MRAG systems towards more imperceptible attacks.
    \item We conduct extensive experiments on two datasets and four versions of retrievers to prove our attack's effectiveness.
\end{itemize}



\section{Related Work}
\label{sec:related_work}
\subsection{Multimodal RAG and Existing Attacks}
Multimodal Retrieval-Augmented Generation~\cite{mei2025mrag-survey, abootorabi2025ask} has emerged as an important technique by extending traditional RAG~\cite{lewis2020rag} to multimodal data, enabling more real-world applications. Despite its huge success, the security issue has gained much attention. Recent studies have highlighted the vulnerabilities of multimodal RAG systems to knowledge poisoning attacks, where malicious information is injected into the external knowledge databases to manipulate the RAG's outputs. MM-Poisoning~\cite{ha2025mmpoison} achieved the attack by constructing query-specific misinformation into injected text and images, or inserting a single irrelevant knowledge instance to fool all queries. Poisoned-MRAG~\cite{liu2025poisoned} formalized the attack as an optimization problem and proposed cross-modal attack strategies to disrupt both retrieval and generation. PoisonedEye~\cite{zhangpoisonedeye} designed injected textual context or optimized the poison image to reduce retrieval performance to achieve attack goals. Despite these advancements, existing research still follows the line of text RAG attacks, which mainly focus on knowledge poisoning attacks. Furthermore, the multimodal characteristics remains underexplored. Thus, in this paper, we propose a new attack on multimodal RAG by only learning small adversarial perturbations added to the image in user query, without modifying the external knowledge base.  

\begin{figure*}[htbp]
    \centering
    \includegraphics[width=1\linewidth]{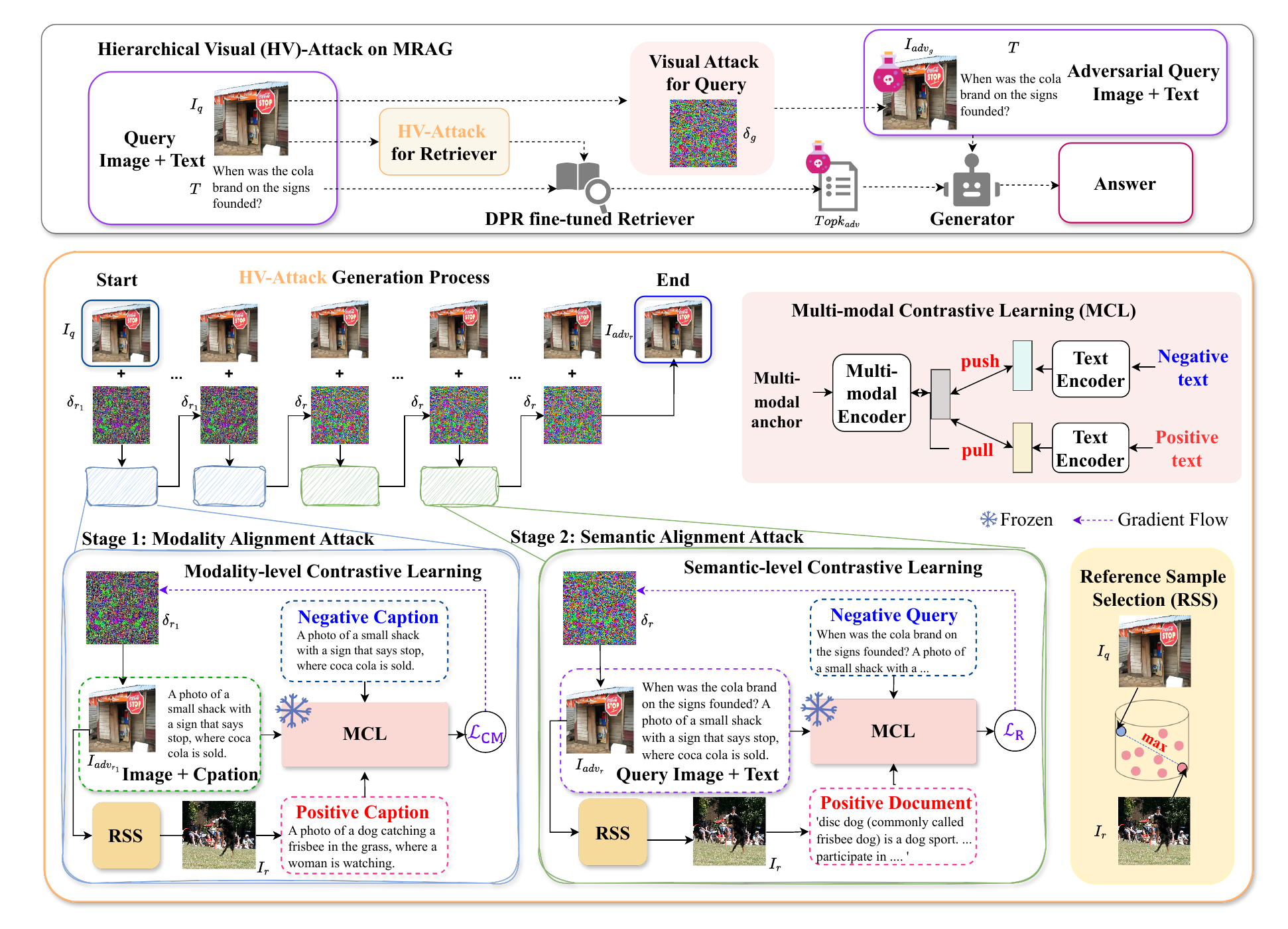}
    \caption{An overview of the proposed method. The top part illustrates the overall MRAG pipeline and our hierarchical structure, which adds image perturbation to the retriever and generator respectively. The hierarchical two-stage strategy we employed for generating retrieval visual attack is shown in the lower block. We optimize the added perturbation step-by-step by first breaking the modality alignment, then disrupting the semantic alignment.}
    \label{fig:method}
\end{figure*}

\subsection{Visual Attack}
Visual adversarial attacks can be categorized into white-box, gray-box and black-box based on the level of knowledge about the attacked model~\cite{zhang2025adversarial}. White-box methods\cite{barach2024cross, wang2024white, guo2024white}, which have full access to the attacked model, aim to maximize the adversarial effect. Fast Gradient Sign Method (FGSM)~\cite{goodfellow2014explaining} generates adversarial examples by adding a small perturbation in the direction of the gradient of the loss function with respect to the input. The Projected Gradient Descent (PGD)~\cite{madry2017towards} is an iterative variant of FGSM that refines the perturbation through multiple steps. The Carlini and Wagner (CW) attack~\cite{carlini2017towards} employs optimization techniques to find adversarial examples that achieve best attack while being close to the original input. Gray-box~\cite{brendel2017decision, ilyas2018black} and black-box~\cite{chen2023rethinking} attacks have limited knowledge about the attacked model, and rely on transferability or surrogate models to craft effective perturbations. In this paper, we focus on white-box attack on images, assuming full access to the open-source retriever in MRAG pipeline.

\section{Method}

\subsection{MRAG Preliminaries and Problem Definition}
A multimodal RAG pipeline consists of a retriever $\mathbf{R}$, a generator $\mathbf{G}$ and an external knowledge base $\mathcal{KB}$. When a user inputs an image $I$ and a text query $T$, the retriever $\mathbf{R}$ first retrieve top-k augmented knowledge from $\mathcal{KB}$. Then the generator $\mathbf{G}$ generates an answer using the multimodal query along with k augmented knowledge. The retrieval process follows the DPR~\cite{karpukhin2020dense} structure, which comprises an image encoder $E_I$ and a text encoder $E_T$. The encoders embed input and knowledge into a common semantic space. A similarity-based retrieval can be conducted by calculating and ranking the distance between the input and knowledge in this semantic space. For each query, the top-k knowledge embeddings with highest similarity are retrieved as results. The multimodal embedding of a user query is obtained by summing the corresponding image and text embeddings, i.e., $E_I(I) + E_T(T)$. The knowledge embedding is depended on its modality. For text corpus, each passage $K_t$'s embedding is $E_T(K_t)$. For multimodal KB, each image-text pair$(K_i, K_t)$'s embedding is $E_I(K_i) + E_T(K_t)$.


In this paper, our goal is to attack the MRAG pipeline by solely modifying the input image. We achieve our attack by misaligning and disrupting the two inputs of the generator compomnent in MRAG, which results in confusion within inputs along with misleading information. To be more specific, for the augmented knowledge, we optimize a small perturbation $\delta_r$ added at the input image $I_q$ to mislead the retriever to return irrelevant knowledge. For the image input of the generator, we add another small perturbation $\delta_g$ that breaks down uni-modal semantic within the image to disrupt image understanding. Consequently, the generator produces incorrect answers based on the misaligned input query and augmented knowledge, with misleading irrelevant contents. The attack objective can be formulated as follows:
\begin{equation*}
\begin{aligned}
    &I_{adv_r} = attack_r(I_q, \delta_r), \,\,\,||I_q-I_{adv_r}||_\infty=||\delta_r||_\infty \, \textless \, \epsilon, \\
    &I_{adv_g} = I_q + \delta_g, \,\,\,||I_q-I_{adv_g}||_\infty=||\delta_g||_\infty \, \textless \, \epsilon, \\
    &Topk_{adv} = \mathbf{R}(T,I_{adv_r},\mathcal{KB}), \\
    &A_{adv} = \mathbf{G}(T, I_{adv_g}, Topk_{adv}),
\end{aligned}
\end{equation*}
where $\delta_r$ and $\delta_g$ are noises added to the image in retrieval and generation, $attack_r(\cdot)$  represents our hierarchical attack method of learning retrieval and generation perturbations respectively, $\epsilon$ is the bound of added perturbation to ensure that it is imperceptible to human eye.

\subsection{Overall Hierarchical Attack Framework}
The overall attack framework is illustrated in \autoref{fig:method}. To produce confusion for the generator in MRAG, we aim to misalign the two inputs of it: the input query and augmented knowledge. Thus, we hierarchically apply two different perturbations to the image, resulting in misleading effect in two directions.

For the augmented knowledge, the added perturbation $\delta_r$ mainly takes effect by disrupting the retrieval process. We further adopt a hierarchical two-stage strategy to learn $\delta_r$, targeting at modality alignment and semantic alignment within retrieval respectively. The detailed design of this hierarchical strategy is described in \autoref{sec:retrieval}. For the generator input query, the added perturbation $\delta_g$ breaks down the uni-modal semantic within the image input, so as to disrupt image understanding in generation.


\subsection{Hierarchical Visual Attack on Augmented Knowledge}
\label{sec:retrieval}
The visual attack on the augmented knowledge as inputs to the generator is achieved by disrupting the retrieval process. We employ a hierarchical two-stage strategy to break down modality and semantic alignments respectively, which are key characteristics in success retrieval.

\textbf{Stage 1: Modality Alignment Attack}

Multimodal dual-encoder retrievers, such as CLIP~\cite{radford2021learning}, achieve strong alignment between text and image modalities. The corresponding text and image embeddings are close in the embedding space, enabling relevant knowledge to be retrieved through the ranking of embedding similarities during cross-modal retrieval. Thus, the first stage of visual attack on augmented knowledge is to break down the multimodal alignment between these modalities.

We achieve this by pushing the query image close to the least similar sample. Given a user query image, we search for a reference image from the database whose embedding has the smallest similarity with the query image’s embedding. Then, given the retriever image encoder $E_i$, text encoder $E_t$ and image captioning model $\mathbf{C}$, we obtain the multimodal image representation $f_{multi_{stage1}}$, the clean query caption's embedding $f_{clean\_cap}$ and the reference image caption's embedding $f_{ref\_cap}$ as follows:
\begin{equation}
    \begin{aligned}
        &f_{clean\_cap} = E_t(\mathbf{C}(I_q)),\\
        &f_{ref\_cap} = E_t(\mathbf{C}(I_r)),\\
        &f_{multi_{stage1}} =  E_i(I_p) + E_t(\mathbf{C}(I_p)),
    \end{aligned}
\end{equation}
where $I_q$ and $I_r$ are the input user query image and its reference image, $I_p$ is the query image with added perturbation. 

To learn the best the perturbation, we conduct contrastive learning between $f_{multi_{stage1}}$, $f_{clean\_cap}$ and $f_{ref\_cap}$ and design the loss function $\mathcal{L}_{CM}$ based on hinge loss~\cite{gentile1998linear}:
\begin{equation}
\begin{aligned}
&\mathcal{L}_{CM} = \max \left( \left\| \text{clean\_sim} - \beta \cdot \text{ref\_sim} \right\| + \gamma, 0 \right), \\
&\text{clean\_sim} = \text{sim}(f_{multi_{stage1}}, f_{clean\_cap}), \\
&\text{ref\_sim} = \text{sim}(f_{multi_{stage1}}, f_{ref\_cap}).
\end{aligned}
\label{eq:lcm}
\end{equation}

As shown in \autoref{eq:lcm}, by minimizing $\mathcal{L}_{CM}$, we minimize the similarity between the query and clean image's caption, while maximize the similarity with the reference caption. The detailed optimization process can be found in \autoref{alg:1}. In this stage, since we focus on cross-modal alignment, we consider the alignment between multimodal representation of the query image and the corresponding text embeddings. Through iterations, we minimize the similarity between the multimodal embedding and its text caption embedding, while maximizing the similarity with the reference text caption embedding. 
\begin{algorithm}[ht]
\caption{Modality Alignment Attack (Stage 1)}
\label{alg:1}
\begin{algorithmic}[1]
\Statex \textbf{Input:} User image $I_q$, reference image $I_r$, retriever image encoder $E_i$, retriever text encoder $E_t$, image captioning model $\mathbf{C}$, generation steps $s$, step length $\alpha$, perturbation bound $\epsilon$, trade-off parameter $\beta$, margin parameter $\gamma$.
\Statex \textbf{Output:} Stage 1's perturbation $\delta_{r_1}$
\State Get clean and reference image caption embedding: $f_{clean\_cap} \gets E_t(\mathbf{C}(I_q))$, $f_{ref\_cap} \gets E_t(\mathbf{C}(I_r))$
\State Initialize perturbation $\delta_{r_1} \gets 0$
\State Initialize perturbed image $I_p \gets I_q + \delta_{r_1}$
\For{$\text{step} \gets 1$ to $s$}
    \State $I_p \gets  I_q + \delta_{r_1}$
    \State $f_{multi_{stage1}} \gets E_i(I_p) + E_t(\mathbf{C}(I_p))$
    \State $\text{clean\_sim} \gets \text{sim}(f_{multi_{stage1}}, f_{clean\_cap})$
    \State $\text{ref\_sim} \gets \text{sim}(f_{multi_{stage1}}, f_{ref\_cap})$
    \State $\mathcal{L}_{CM} \gets\max \left( \left\| \text{clean\_sim} - \beta \cdot \text{ref\_sim} \right\| + \gamma, 0 \right),$
    \State Optimize $\delta_{r_1} \gets \delta_{r_1} - \alpha \cdot \text{sign}(\nabla_{\delta_{r_1}} \mathcal{L}_{CM})$
    \State $\delta_{r_1} \gets \text{Clip}(\delta_{r_1}, -\epsilon, \epsilon)$
\EndFor
\Statex \textbf{Return} $\delta_{r_1}$
\end{algorithmic}
\end{algorithm}

\textbf{Stage 2: Semantic Alignment Attack}

The second stage of our attack aims to further break down the semantic relevance between cross-modal embeddings, which is crucial for retrieving useful knowledge for generation.
The form of the loss function is similar to that in the first stage. However, the contrastive learning is now conducted between the multimodal representation of user query, the query text embedding and reference passage embedding. We retrieve the positive text knowledge for the reference image and use its embedding as the positive embedding in this stage. The embeddings are obtained by:
\begin{equation}
    \begin{aligned}
       & f_{clean\_query} = E_t(T_q + \mathbf{C}(I_q)), \\
       &f_{ref\_passage} = E_t(T_p), \\
       &f_{multi_{stage2}} = E_i(I_p) + E_t(T_q + \mathbf{C}(I_p)),
    \end{aligned}
\end{equation}
where $T_q$ and $T_p$ are the user text query and retrieved reference text knowledge respectively.

The contrastive learning of the second stage is designed as shown in \autoref{eq:lr}:
\begin{equation}
    \begin{aligned}
&\mathcal{L}_R \gets\max \left( \left\| \text{clean\_sim} - \beta \cdot \text{ref\_sim} \right\| + \gamma, 0 \right), \\
&\text{clean\_sim} \gets \text{sim}(f_{multi_{stage2}}, f_{clean\_query}), \\
&\text{ref\_sim} \gets \text{sim}(f_{multi_{stage2}}, f_{ref\_passage}). 
    \end{aligned}
    \label{eq:lr}
\end{equation}

The detailed process of the second stage is shown in \autoref{alg:2}. Building on the perturbation generated in the first stage, the second stage iteratively refines and outputs the final perturbation.
\begin{algorithm}[htbp]
\caption{Semantic Alignment Attack (Stage 2)}
\label{alg:2}
\begin{algorithmic}[1]
\Statex \textbf{Input:} Perturbation $\delta_{r_1}$ from first stage, user image $I_q$, User text query $T_q$, reference positive passage $T_p$, models $E_i, E_t, \mathbf{C}$ and hyperparameters $s, \alpha, \epsilon, \beta, \gamma$ the same as stage 1.

\Statex \textbf{Output:} Final Perturbation $\delta_r$
\State Get reference positive passage embedding: $f_{ref\_passage} \gets E_t(T_p)$
\State Get clean query text embedding: $f_{clean\_query} \gets E_t(T_q + \mathbf{C}(I_q))$
\State Initialize perturbation $\delta_r \gets \delta_{r_1}$
\State Initialize perturbed image $I_p \gets I_q + \delta_r$
\For{$\text{step} \gets 1$ to $s$}
    \State $I_p \gets  I_q + \delta_r$
    \State $f_{multi_{stage2}} \gets E_i(I_p) + E_t(T_q + \mathbf{C}(I_p))$
    \State $\text{clean\_sim} \gets \text{sim}(f_{multi_{stage2}}, f_{clean\_query})$
    \State $\text{ref\_sim} \gets \text{sim}(f_{multi_{stage2}}, f_{ref\_passage})$
    \State $\mathcal{L}_R \gets\max \left( \left\| \text{clean\_sim} - \beta \cdot \text{ref\_sim} \right\| + \gamma, 0 \right),$
    \State Optimize $\delta_r \gets \delta_r - \alpha \cdot \text{sign}(\nabla_{\delta_r}  \mathcal{L}_R)$
    \State $\delta_r \gets \text{Clip}(\delta_r, -\epsilon, \epsilon)$
\EndFor
\Statex \textbf{Return} $\delta_r$
\end{algorithmic}
\end{algorithm}

With the two-stage hierarchical strategy, we obtain the final retrieval perturbation added to the query image and retrieve the adversarial augmented knowledge. The generator's disrupted query is obtained by adding noise generated by advanced attack method on LMMs to the image. The two inputs that misalign with each other achieve the hierarchical attack on MRAG pipeline.
\section{Experiments}
\begin{table*}[htbp]\small
\centering
\caption{VQA Performance on OK-VQA. (ASR* is calculated as $(s_{clean}-s_{adv})/s_{clean}$ for each metric $s$, EM: Exact Match,  “↑” indicates that a higher value is better for this metric, while “↓” indicates that a lower value is better. All VQA metrics are reported with RAG knowledge number $K=5$. \textbf{Bold} and \underline{underline} represent the best and second best results respectively.)}
\label{tab:vqa-okvqa}
\begin{tabular}{c| l| cc|cc| cc|cc }
\toprule
\multirow{3}{*}{\textbf{Retriever}} & \multirow{3}{*}{\textbf{Method}} & \multicolumn{4}{c|}{\textbf{Retrieval(\%)}}
&\multicolumn{4}{c}{\textbf{\textbf{BLIP-2 VQA}(\%)}} \\

\cmidrule(r){3-10}
 &  & \textbf{R@5(↓)}&\textbf{ASR*(↑)} & \textbf{P@5(↓)} &\textbf{ASR*(↑)}& \makecell[c]{\textbf{VQA}\\ \textbf{Score(↓)}} & \textbf{ASR*(↑)} & \textbf{EM(↓)} & \textbf{ASR*(↑)}  \\

\midrule
\multirow{5}{*}{\shortstack{CLIP \\ ViT-L/14}} 
& Clean  & 50.57 & - & 27.86 &- & 38.53& - &41.76& - \\
\cmidrule(r){2-10}
&AnyAttack~\cite{zhang2025anyattack} & 50.12& 0.89& 27.21& 2.33 & 37.94& 0.59& 41.14& 0.62\\ 

&X-Transfer~\cite{huang2025xtransfer} & 48.14& 4.81& 25.27&9.30& 32.95& 14.48& 36.03&13.72\\

&LMM Attack~\cite{cui2024robustness} & \textbf{12.76}& \textbf{74.77}& \textbf{4.48}& \textbf{83.92} & \underline{32.53} & \underline{15.57} & \underline{35.26}& \underline{15.57}\\



&Ours & \underline{14.00} & \underline{72.32}& \underline{4.86}& \underline{82.56}& \textbf{28.39}&\textbf{26.32}&\textbf{30.99}&\textbf{25.79}\\

\midrule
\multirow{5}{*}{\shortstack{CLIP \\ViT-L/14 \\ \textit{finetuned}}} & Clean & 74.63& -& 44.03& - &41.25& - & 44.55& - \\
\cmidrule(r){2-10}
&AnyAttack~\cite{zhang2025anyattack} & 74.18& 0.60& 43.56& 1.07 &41.11& 0.34& 44.47&0.18\\
&X-Transfer~\cite{huang2025xtransfer} & 72.69& 2.60& 40.99& 6.90&35.54 & 13.84& 38.55&13.47\\
&LMM Attack~\cite{cui2024robustness} & \underline{56.92}&\underline{23.73} & \underline{27.34}&  \underline{37.91}&  \underline{33.38}& \underline{19.08} &\underline{36.29} & \underline{18.54}\\


&Ours & \textbf{31.81} & \textbf{57.38}& \textbf{12.40}& \textbf{71.84}& \textbf{28.80}& \textbf{30.18}&\textbf{31.25}&\textbf{29.85}\\

\bottomrule
\end{tabular}
\end{table*}

\subsection{Datasets and evaluation metrics}
We conduct our experiments on two wide-used datasets:
\begin{itemize}
    \item \textbf{OK-VQA}~\cite{marino2019ok}: This is a large knowledge-based VQA dataset. Each data sample consists of a question, an image and 10 gold answers. The images are from the COCO dataset~\cite{lin2014microsoft} and each question requires external knowledge to answer. We use the test set for attack evaluation, which contains 5046 samples. For OK-VQA, we follow RAVQA~\cite{lin2022retrieval}, FLMR~\cite{lin2023fine} to use the Google Search corpus as the external knowledge base, which contains 169,306 passages in total.
    \item \textbf{Infoseek}~\cite{chen2023can}: We follow PoisonedEye~\cite{zhangpoisonedeye} to attack 1000 samples from the visual question answering dataset Infoseek for evaluation. The knowledge database is the same subset of 2M image-text pairs from OVEN-Wiki~\cite{hu2023open} with PoisonedEye.
\end{itemize}

\begin{table*}[h]\small
\centering
\caption{VQA Performance on InfoSeek. (ASR* is calculated as $(s_{clean}-s_{adv})/s_{clean}$ for each metric $s$, EM: Exact Match,  “↑” indicates that a higher value is better for this metric, while “↓” indicates that a lower value is better. All VQA metrics are reported with RAG knowledge number $K=5$. \textbf{Bold} and \underline{underline} represent the best and second best results respectively.)}
\label{tab:vqa-infoseek}
\begin{tabular}{c| l| cc|cc| cc |cc}
\toprule
\multirow{3}{*}{\textbf{Retriever}} & \multirow{3}{*}{\textbf{Method}} & \multicolumn{4}{c|}{\textbf{Retrieval(\%)}}
&\multicolumn{2}{c|}{\textbf{\textbf{BLIP-2 VQA}(\%)}} &\multicolumn{2}{c}{\textbf{\textbf{LLava VQA}(\%)}}\\

\cmidrule(r){3-10}
 &  & \textbf{R@5(↓)}&\textbf{ASR*(↑)} & \textbf{P@5(↓)} &\textbf{ASR*(↑)}& \textbf{EM} & \textbf{ASR*(↑)}& \textbf{EM} & \textbf{ASR*(↑)}\\

\midrule
\multirow{4}{*}{Siglip-so400m} 
& Clean  & 44.47 & - & 20.12 &- & 21.63&-&31.96 & - \\
\cmidrule(r){2-10}
&X-Transfer~\cite{huang2025xtransfer} & 24.42& 45.09&11.01 & 45.28&  \underline{11.79}& \underline{45.49}& 19.93& 37.64\\
&LMM Attack~\cite{cui2024robustness} & \underline{5.35}& \underline{87.97}& \underline{1.68}&  \underline{91.65}&11.91& 44.94&\textbf{11.79}&\textbf{63.11} \\
&Ours & \textbf{4.37}&\textbf{90.17} & \textbf{1.56}& \textbf{92.25}& \textbf{5.47}  &\textbf{74.71}& \underline{12.52}&\underline{60.83} \\
\midrule
\multirow{4}{*}{CLIP ViT-H} & Clean & 43.26& -& 18.78& - & 20.41& - &29.77 & -\\
\cmidrule(r){2-10}
&AnyAttack~\cite{zhang2025anyattack} & 32.68& 24.46& 14.09&  24.97& 16.16&20.82 &22.96&22.88\\
&X-Transfer~\cite{huang2025xtransfer} & 13.49& 68.82& 4.45& 76.30& \underline{5.10}&\underline{75.01}&11.91 & 59.99\\
&LMM Attack~\cite{cui2024robustness} & \underline{4.86}& \underline{88.77}& \underline{1.43}&  \underline{92.39}&  13.00& 36.31 & \underline{9.48}&\underline{68.16}\\
&Ours &\textbf{2.31}  & \textbf{94.66}&\textbf{0.51}& \textbf{97.28}& \textbf{3.28}&\textbf{83.93} & \textbf{10.45}&\textbf{64.90}\\

\bottomrule
\end{tabular}
\end{table*}

\textbf{Evaluation metrics.} Our attack method is evaluated from two perspectives: retrieval and generation. For retrieval, we first use $S(q, p)$ to identify the relation between a query $q$ and a passage $p$, which is classified based on whether the passage contains a ground-truth answer to the query.
\begin{equation*}
S(q, p)=\left\{
	\begin{aligned}
	&1, \quad \mbox{if p contains answer to q},\\
	&0, \quad \mbox{if p does not contain answer to q}.\\
	\end{aligned}
	\right
	.
\label{eq:relation}
\end{equation*}
Then, we adopt the following two retrieval metrics.
\begin{itemize}
    \item \textbf{Recall@K} evaluates how likely the retrieved K passages are to contain answers to the query, which is the proportion of queries that have positive passages in the retrieved results:
        \begin{equation}
            \label{eq:recall}
            Recall@K=\min (\sum_{k=1}^KS(q, p_k), 1).
        \end{equation}
    \item \textbf{Precision@K} evaluates how much percent of the retrieved K passages contain answers to the query:
        \begin{equation}
            \label{eq:precision}
            Precision@K=\frac{1}{K}\sum_{k=1}^KS(q, p_k).
        \end{equation}
\end{itemize}

For generation, we adopt the corresponding metric for each dataset. For OK-VQA, VQA score and Exact Match are used following RAVQA~\cite{lin2022retrieval}. For InfoSeek, Exact Match is calculated. For each metric, we calculate the non-target attack success rate proposed by X-transfer~\cite{huang2025xtransfer}, which is $(s_{clean}-s_{adv})/s_{clean}$ for each metric $s$.

\subsection{Implementation Details}
We implement the proposed method based on the open-source PyTorch~\cite{paszke2019pytorch} framework. All the experiments are conducted on NVIDIA RTX3090. We adopt PGD-step 50, step size $\alpha=1/255$, perturbation bound $\epsilon =8/255$ for the two stages respectively in retrieval perturbation generation. The trade-off parameter $\beta$ and margin parameter $\gamma$ are set to 0.4 and 0.6 in all loss functions. We adopt the X-transfer noise as $\delta_g$ in our experiments.

For OK-VQA, we use the off-the-shell CLIP ViT-L/14 as well as a fine-tuned version of it as retrievers. For Infoseek, we use the off-the-shell Siglip-so400m and CLIP ViT-H as retrievers following PoisonedEye~\cite{zhangpoisonedeye}. For generators, we use the off-the-shell BLIP-2-flan-T5-xl and LLava-NEXT models.

\subsection{Baseline Models}
We compare our method with various state-of-the-art visual noise generation algorithms. AnyAttack~\cite{zhang2025anyattack} finetuned a decoder to generate noise for any image that transforms it to any target. We use the released AnyAttack decoder to generate noises using the reference image in our method for comparison. X-transfer~\cite{huang2025xtransfer} learned a transferrable noise that can be applied to any image. We employ the "xtransfer\_large\_linf\_eps12\_non\_targeted" noise and set $\epsilon = 8$ to apply to all the images. LMM Attack~\cite{cui2024robustness} learned the noise using PGD algorithm, with the cross entropy loss to minimize similarity between the adversarial image and its image caption as the objective function. Since the code was not released, we re-implemented the algorithm and set PGD-step to 50.

\subsection{Experimental Results}
\subsubsection{Main Attack Results}
The main attack results of attacking the MRAG system on two datasets are shown in \autoref{tab:vqa-okvqa} and \autoref{tab:vqa-infoseek}. The overall results show that our proposed method achieved the most severe attack influence on retrieval and generation process of MRAG. Several key conclusions can be made:

\textbf{Our hierarchical attack works on both retrieval and generation.} Among the datasets, our hierarchical attack method all achieve declines in both retrieval and generation metrics. As shown in the results, former baseline methods demonstrate attack advantage either in retrieval or generation. X-transfer~\cite{huang2025xtransfer}'s noise is more effective at attacking the generator (the retrieval metrics' decline are subtle compared to the VQA metrics). While LMM attack~\cite{cui2024robustness}'s noise is comparably more effective at attacking the retrieval process. For our attack, the hierarchical structure leads to a more balanced attack effectiveness on both retrieval and generation, and disrupts the whole MRAG chain by only modifying the image input, making the attack highly imperceptible to human vision while effective.

\textbf{Our hierarchical attack works on the fine-tuned retriever.} As shown in the tables, the off-the-shell models are highly vulnerable against adversarial noise in the retrieval task. However, previous attack methods can not achieve equivalent effectiveness towards both off-the-shell and fine-tuned models. For example, as shown in \autoref{tab:vqa-okvqa}, the non-hierarchical LMM attack can achieve 74.77\% attack success rate on the R@5 metric of off-the-shell CLIP, while only getting 23.73\% success rate on the fine-tuned version. While for our hierarchical attack method, we achieved high success rate (72.32\% and 57.38\%) on non-fine-tuned and fine-tuned retrievers. Note that though the attack performance of LMM Attack and our method are compatible on the original CLIP, our method achieves over two times the success rate on the fine-tuned version.

\textbf{Our hierarchical attack works on various black-box generators.} We use BLIP-2 and LLaVA as black-box generators. VQA results of OK-VQA using LLaVA is shown in \autoref{tab:vqa-llava}. With using different black-box generators, our attack with adversarial augmented knowledge and query noise can all achieve damaging attack effect.

\begin{table}[htbp]\small
    \centering
    \begin{tabular}{c|cc|cc}
    \toprule
      &  \makecell[c]{\textbf{VQA}\\ \textbf{Score(↓)}}  & \textbf{ASR*(↑)} & \textbf{EM(↓)} & \textbf{ASR*(↑)}\\
      \midrule
       Clean  & 63.30& -&67.34 &-\\
       \midrule
       AnyAttack & 61.31& 3.14&65.46& 2.79\\
       X-transfer& 57.17& 9.68&60.98 & 9.44\\
       LMM Attack& \underline{56.72} & \underline{10.39}& \underline{60.42}&\underline{10.23}\\
       Ours &\textbf{54.19} & \textbf{14.39}&\textbf{57.77}&\textbf{14.21} \\
       \bottomrule
    \end{tabular}
    \caption{VQA results on OK-VQA with LLaVA as generator. Top-5 retrieved documents are used for generation.}
    \label{tab:vqa-llava}
\end{table}


\subsubsection{Ablation Studies}
\textbf{Analysis of Hierarchical Retrieval Attack.} \autoref{tab:retrieval-okvqa} and \autoref{tab:retrieval-infoseek} show the detailed retrieval results on OK-VQA and Infoseek datasets. Both stages of the attack demonstrate individual effectiveness. Using stage 1 and stage 2 alone both degrades retrieval metrics. Stage 2's individual attack performance is comparably better than stage 1, since its attack goal is more directly targeted at retrieval. While across all datasets and retriever evaluated, the hierarchical two-stage attack consistently achieves optimal performance, demonstrating the attack effect of breaking both modality and semantic alignment.





\begin{table}[htbp]\small
\centering
\caption{Retrieval Performance on OK-VQA. (\textbf{Bold} and \underline{underline} represent the best and second best results respectively.)}
\label{tab:retrieval-okvqa}
\begin{tabular}{c|cccc}
\toprule
 \multicolumn{5}{c}{CLIP ViT-L/14} \\
\midrule
 &\textbf{R@5}& \textbf{R@10} & \textbf{P@5} &\textbf{P@10} \\
\midrule
Clean & 50.57 & 62.25 & 27.86 &27.48 \\
\midrule
AnyAttack~\cite{zhang2025anyattack} & 50.12 & 61.06 & 27.21 & 26.86\\ 
X-Transfer~\cite{huang2025xtransfer} & 48.14 & 59.02& 25.27 & 24.73\\
LMM Attack~\cite{cui2024robustness} &\textbf{12.76} & \textbf{19.26}& \textbf{4.48} & \textbf{4.65}\\
\midrule
Ours \textit{w/ }Stage 1 &21.94 & 31.85&8.52 & 8.74\\
Ours \textit{w/ }Stage 2 &19.70 &28.22 &7.56 & 7.77\\
Ours & \underline{14.00} & \underline{20.77}& \underline{4.86} &\underline{4.93} \\
\midrule

\multicolumn{5}{c}{CLIP ViT-L/14 \textit{finetuned}}\\
\midrule
& \textbf{R@5} & \textbf{R@10} & \textbf{P@5} &\textbf{P@10} \\
\midrule
Clean & 74.63 & 83.31 & 44.03 & 42.27 \\
\midrule
AnyAttack~\cite{zhang2025anyattack} & 74.18& 82.82& 43.56 & 41.83 \\ 
X-Transfer~\cite{huang2025xtransfer} & 72.69 & 81.25& 40.99 & 38.91\\
LMM Attack~\cite{cui2024robustness} & 56.92& 68.59& 27.34&26.56 \\ 
\midrule
Ours \textit{w/ }Stage 1 &54.48 & 65.93& 25.73 & 25.07 \\ 
Ours \textit{w/ }Stage 2 & \underline{37.65}& \underline{49.68}& \underline{15.56} &\underline{15.78} \\
Ours & \textbf{31.81}& \textbf{44.19}& \textbf{12.40} &\textbf{12.78} \\ 

\bottomrule
\end{tabular}
\end{table}

\begin{table}[htbp]\small
\centering
\caption{Retrieval Performance on Infoseek. (\textbf{Bold} and \underline{underline} represent the best and second best results respectively.)}
\label{tab:retrieval-infoseek}
\begin{tabular}{c| cccc}
\toprule
 \multicolumn{5}{c}{Siglip-so400m} \\
\midrule
& \textbf{R@5} & \textbf{R@10} & \textbf{P@5} & \textbf{P@10} \\
\midrule
Clean & 43.47 &51.89 &19.61 &17.06\\
\midrule
X-Transfer~\cite{huang2025xtransfer} & 24.42 &30.01& 11.01 &9.68\\
LMM Attack~\cite{cui2024robustness} & 5.35& 8.14& 1.68& 1.7\\
\midrule
Ours \textit{w/ }Stage 1 &5.1 &8.51 &1.97 &1.91 \\
Ours \textit{w/ }Stage 2&\underline{4.74} & \underline{7.9} & \textbf{1.51} & \underline{1.51}\\
Ours & \textbf{4.37} & \textbf{7.29} & \underline{1.56} & \textbf{1.51}\\
\midrule

\multicolumn{5}{c}{CLIP ViT-H}\\
\midrule
& \textbf{R@5} & \textbf{R@10} & \textbf{P@5} & \textbf{P@10} \\
\midrule
Clean & 43.26 & 48.97 & 18.78 & 15.55\\
\midrule
AnyAttack~\cite{zhang2025anyattack} &  32.69 & 40.34 & 14.09 & 11.85\\ 
X-Transfer~\cite{huang2025xtransfer} & 13.49 & 18.59 & 4.45 & 3.94\\
LMM Attack~\cite{cui2024robustness} & 4.86 & 7.65 & 1.43 & 1.49 \\ 
\midrule
Ours \textit{w/ }Stage 1 &3.77 & 5.95 & 0.92 & 0.89 \\ 
Ours \textit{w/ }Stage 2 & \underline{2.61} & \underline{4.39} & \underline{0.69} & \underline{0.69}\\
Ours & \textbf{2.31} & \textbf{3.89} & \textbf{0.51} & \textbf{0.52} \\ 

\bottomrule
\end{tabular}
\end{table}



\begin{figure}
    \centering
    \includegraphics[width=1\linewidth]{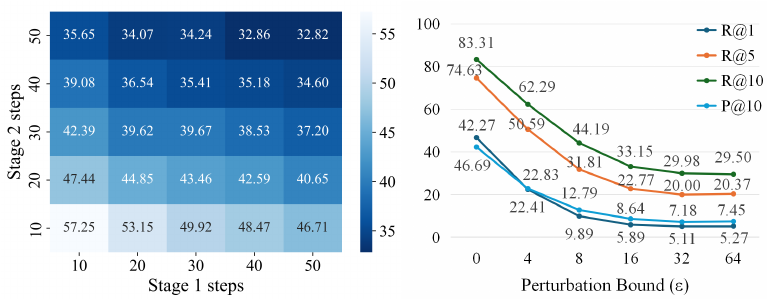}
    \caption{Performance of ablated models with different steps.}
    \label{fig:steps}
\end{figure}
\textbf{Effect of PGD Steps.} As shown in \autoref{fig:steps} (left), we report the Recall@5 metric on OK-VQA dataset with fine-tuned CLIP using 10 to 50 steps of stage 1 and 2. As the PGD algorithm increases optimization steps, the attack results become better, eventually coming to convergence. We finally choose PGD-step50 for each stage of optimization.

\textbf{Effect of Perturbation Budget.} We explore the effect of setting $\epsilon$ to different values. As shown in \autoref{fig:steps} (right), when $\epsilon$ gets larger, the attack effect is more obvious. Different colors refer to various retrieval metrics, when $\epsilon$ increases, all the metrics degrade. With a limited budget $\epsilon = 8$, attack effectiveness can already be achieved while being imperceptible.

\subsubsection{Case studies}
\begin{figure}
    \centering
    \includegraphics[width=1\linewidth]{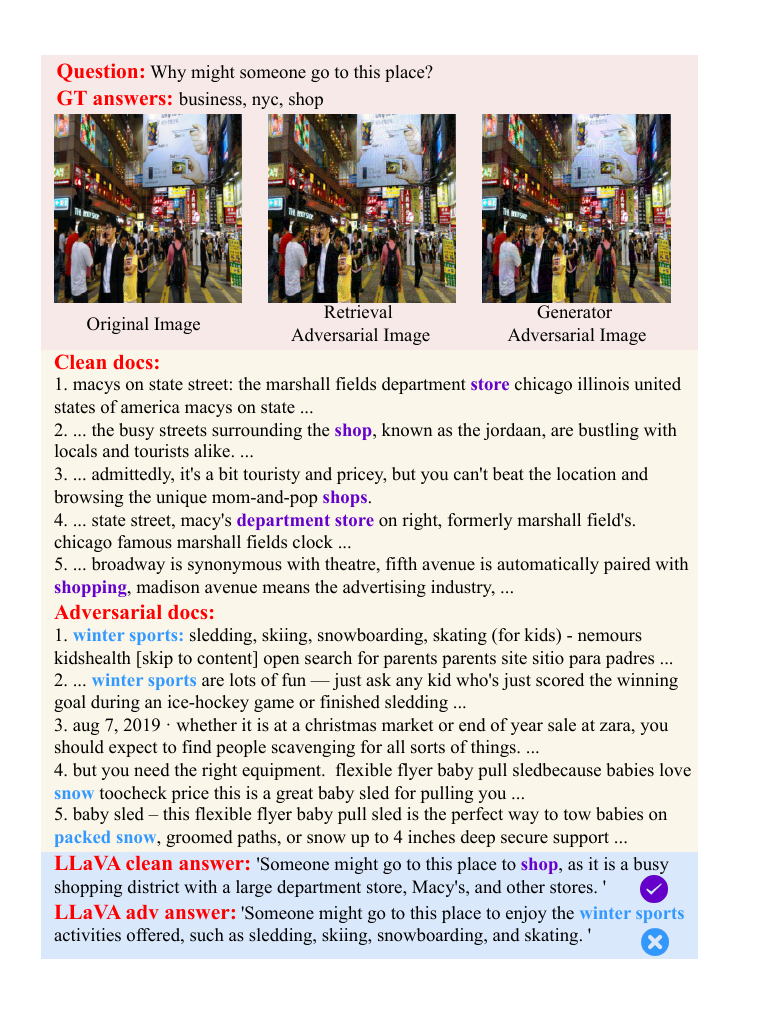}
    \caption{An example showing the original image, the adversarial images as inputs to retriever and generator, as well as the clean and adversarial augmented knowledge, along with the generated answer based on them. The example demonstrates the attack effect of our hierarchical method within imperceptible disruption.}
    \label{fig:case}
\end{figure}

We provide a case study in \autoref{fig:case}. This example shows the attack effect of our hierarchical method under the perturbation budget $\epsilon=8$. As shown in the example, with the original query image without noise, the retriever returns augmented knowledge that are closely related to answering the question. The generator provides a correct answer accordingly. However, with added imperceptible noise, the retriever returns irrelevant augmented knowledge that are about winter sports. With the disruptions, the final prediction of the generator is wrong and unreasonable.  

\section{Conclusion}
In this paper, we proposed a novel attack method against MRAG pipeline that focused solely on image inputs. With hierarchical optimization, we target MRAG's retriever and generator across different levels of abstraction, achieving severe while stealth attack impact. Our research reveals that MRAG technologies, while widely adopted in practice, remain vulnerable to security threats posed by imperceptible adversarial visual noise. Future work will focus on uncovering deeper and more diverse potential threats brought by visual attacks on MRAG systems and developing robust defense mechanisms to balance effectiveness and security.
{
    \small
    \bibliographystyle{ieeenat_fullname}
    \bibliography{main}
}


\end{document}